\documentclass{article}
\usepackage{ijcai20}

\usepackage{times}
\usepackage{soul}
\usepackage{url}
\usepackage[hidelinks]{hyperref}
\usepackage[utf8]{inputenc}
\usepackage[small]{caption}
\usepackage{graphicx}
\usepackage{amsmath}
\usepackage{amsthm}
\usepackage{booktabs}
\usepackage{algorithm}
\usepackage{algorithmic}
\usepackage{subcaption}
\usepackage{amsfonts}
\usepackage{diagbox}
\usepackage{textcomp}
\title{Task-Level Curriculum Learning for Non-Autoregressive Neural Machine Translation}

\author{
Jinglin Liu$^1$\thanks{Equal contribution.}\and
Yi Ren$^1$\footnotemark[1]\and
Xu Tan$^2$\and
Chen Zhang$^1$\and
Tao Qin$^2$\and
Zhou Zhao$^1$\footnote{Corresponding author}\And
Tie-Yan Liu$^2$\\
\affiliations
$^1$Zhejiang University\\
$^2$Microsoft Research Asia\\
\emails
\{jinglinliu,rayeren,zc99,zhaozhou\}@zju.edu.cn,\\
\{xuta,taoqin,tyliu\}@microsoft.com
}

\begin{document}

\maketitle

\begin{abstract}
Non-autoregressive translation (NAT) achieves faster inference speed but at the cost of worse accuracy compared with autoregressive translation (AT). Since AT and NAT can share model structure and AT is an easier task than NAT due to the explicit dependency on previous target-side tokens, a natural idea is to gradually shift the model training from the easier AT task to the harder NAT task. To smooth the shift from AT training to NAT training, in this paper, we introduce semi-autoregressive translation (SAT) as intermediate tasks. SAT contains a hyperparameter $k$, and each $k$ value defines a SAT task with different degrees of parallelism. Specially, SAT covers AT and NAT as its special cases: it reduces to AT when $k=1$ and to NAT when $k=N$ ($N$ is the length of target sentence). We design curriculum schedules to gradually shift $k$ from $1$ to $N$, with different pacing functions and number of tasks trained at the same time. We called our method as task-level curriculum learning for NAT (TCL-NAT). Experiments on IWSLT14 De-En, IWSLT16 En-De, WMT14 En-De and De-En datasets show that TCL-NAT achieves significant accuracy improvements over previous NAT baselines and reduces the performance gap between NAT and AT models to 1-2 BLEU points, demonstrating the effectiveness of our proposed method.

\end{abstract}

\section{Introduction}
Neural Machine Translation (NMT) has witnessed rapid progress in recent years~\cite{DBLP:journals/corr/BahdanauCB14,DBLP:conf/icml/GehringAGYD17,DBLP:conf/nips/VaswaniSPUJGKP17}. Typically, NMT models adopt the encoder-decoder framework~\cite{DBLP:journals/corr/BahdanauCB14}, and the decoder generates a target sentence in an autoregressive manner~\cite{DBLP:journals/corr/BahdanauCB14,DBLP:conf/nips/VaswaniSPUJGKP17}, where the generation of the current token depends on previous tokens and the source context from the encoder. While the accuracy of NMT models achieve human parity, they usually suffer from high inference latency due to autoregressive generation. Therefore, non-autoregressive translation (NAT)~\cite{gu2017non,guo2019non,wang2019non,ma2019flowseq,ren2019fastspeech} has been proposed to generate target tokens in parallel, which can greatly speed up the inference process. 

However, the accuracy of NAT models still lag behind that of the autoregressive translation (AT) models, due to the previous target tokens are removed from conditional dependency. A variety of works have tried to improve the accuracy of NAT, including enhanced decoder input with embedding mapping~\cite{guo2019non}, generative flow~\cite{ma2019flowseq}, and iterative refinement~\cite{ghazvininejad2019mask,lee2018deterministic}, etc. However, none of these works leverage the task relationship between AT and NAT when designing their methods. As AT models are more accurate and easier to train than NAT models due to the explicit dependency on previous tokens, a natural idea is to first train the model with easier AT, and then continue to train it with harder NAT.

AT and NAT can be regarded as two tasks that are far different from each other, which makes it less beneficial to directly shift to NAT training right after AT training. How to smoothly shift the model training from AT to NAT is critical for the final accuracy. In this paper, we introduce semi-autoregressive translation (SAT)~\cite{wang2018semi}, which only generates a part of the tokens in parallel at each decoding step, as intermediate tasks to bridge the shift process from AT to NAT. Specifically, we define a parameter $k$ to represent the degree of parallelism for each task, and view different tasks under a unified perspective: $k=1$ represents AT, $k=N$ represents NAT where $N$ is the length of target sentence, and $1<k<N$ represents SAT. Intuitively, a task with smaller $k$ is easier to train and achieves higher accuracy, while that with larger $k$ is harder to train and results in worse accuracy~\cite{wang2018semi}, which forms a good curriculum to train the model from easy to hard.

Inspired by this, we propose a task-level curriculum learning for non-autoregressive translation (TCL-NAT), which trains the model with sequentially increased $k$. We divide the training procedure into three phases: AT training ($k=1$), SAT training ($1<k<N$) and NAT training ($k=N$). SAT training consists of multiple stages, where we shift $k$ gradually and exponentially as $k=2, 4, 8, ..., 16$. To find the best schedule strategy to shift $k$, we design different pacing functions to control the training steps for each $k$, including linear, logarithmic and exponential functions. On the other hand, to smooth the shift process and reduce the gap between different stages, we further introduce a parameter called task window $w$, which represents the number of tasks training at the same time in each stage. For example, when $w=2$, we train the model with $k=1, 2$ for the first stage and $k=2, 4$ for the second stage, and so on.

We implement TCL-NAT on Transformer model~\cite{DBLP:conf/nips/VaswaniSPUJGKP17}. In order to support different $k$ in the same model, we introduce a causal-$k$ self-attention mechanism in the Transformer decoder. We conduct experiments on four translation datasets including IWSLT14 German-English (De-En), IWSLT16 English-German (En-De), WMT14 English-German (En-De) and WMT14 German-English (De-En) to demonstrate the effectiveness of our method. The experiment results show that our method can achieve significant improvement over NAT baselines and also outperform state of the art NAT models, without sacrificing the inference speed. Specifically, we outperform the state of art NAT model \cite{guo2019fine} by 1.88 BLEU on the IWSLT14 De-En task, and reduce the accuracy gap between AT and NAT models to nearly 1 BLEU point on IWSLT16 En-De and WMT14 En-De tasks.

\section{Related Work}
In this section, we first introduce the related works on neural machine translation, including autoregressive translation (AT), non-autoregressive translation (NAT) and semi-autoregressive translation (SAT), and then describe three learning paradigms: transfer learning, multitask learning and curriculum learning, which are related to our method. 

\subsection{Neural Machine Translation (AT/NAT/SAT)}
An autoregressive translation (AT) model takes source sentence $s$ as input and then generates the tokens of target sentence $y$ one by one during the inference process~\cite{DBLP:journals/corr/BahdanauCB14,sutskever2014sequence,DBLP:conf/nips/VaswaniSPUJGKP17}, which causes much inference latency. To improve the inference speed of AT models, a series of works develop non-autoregressive translation (NAT) models based on Transformer~\cite{gu2017non,lee2018deterministic,li2019hint,wang2019non,guo2019non}, which generate all the target tokens in parallel. Several works introduce auxiliary components or losses to improve the accuracy of NAT models: Wang et al. \shortcite{wang2019non} and Li et al. \shortcite{li2019hint} propose auxiliary loss functions to solve the problem that NAT models tend to translate missing and duplicating tokens; Guo et al. \shortcite{guo2019non} try to enhance the decoder input with target-side information by leveraging auxiliary information; Ma et al. \shortcite{ma2019flowseq} introduce generative flow to directly model the joint distribution of all target tokens simultaneously. While NAT models achieve faster inference speed, the translation accuracy is still worse than AT model. Some works aim to balance the translation accuracy and inference latency between AT and NAT by introducing semi-autoregressive translation (SAT)~\cite{wang2018semi}, which generates multiple adjacent tokens in parallel during the autoregressive generation.

%The conditional probability can be defined as $P(y|x)=\prod\limits_{t=1}^{N}P(y_t|x)$, where $N$ is the length of the target tokens. 
% The fertility prediction function varies among works, but the common idea is finding a  proper way to obtain the decoder input from the source sentence in the shortest time. The typical designs, for example, \cite{gu2017non} adds fertility module on each encoder output to predict the replication times of source tokens sending to decoder as input;~\cite{ren2019fastspeech} expands the encoder's hidden sequence and sends that to decoder without any computing;~\cite{guo2019non} designs a mapping mechanism using additional information to enhance the decoder input generated from source token. \cite{lee2018deterministic} proposed an iterative refinement policy to estimate optimal decoder input and $F_y$. 

Different from the above works, we leverage AT, SAT and NAT together and schedule the training in a curriculum way to achieve better translation accuracy for NAT.

\subsection{Transfer/Multitask/Curriculum Learning}
Our proposed TCL-NAT actually leverages the knowledge from easier and more accurate tasks $k<N$ to help the task $k=N$, but uses a curriculum schedule and trains multiple tasks at a stage. In general, our work is related to three different learning paradigms: transfer learning, multitask learning and curriculum learning.  

Transfer learning has been a common approach for NLP tasks. Pre-trained models such as BERT \cite{devlin2018bert} and MASS \cite{song2019mass} are fine-tuned on many language understanding and generation tasks for better accuracy. Many NAT works~\cite{gu2017non,lee2018deterministic,guo2019non,guo2019fine} employ sequence level data distillation to transfer the knowledge from AT teacher model to NAT student model which has proved to be effective. 
%Li et al. \shortcite{li2019hint} add constraints in hidden layers and word alignments during training NAT model with hints derived from a well-trained AT model with the same architecture, also achieve improvement over previous works

Multitask learning \cite{caruana1997multitask} has found extensive usage in NLP tasks. Dong et al. \shortcite{dong2015multi} use multitask learning for multiple language translation. Anastasopoulos and Chiang \shortcite{anastasopoulos2018tied} explore multitask models for neural translation of speech and find that jointly trained models improve performance on the tasks of low-resource speech transcription and translation. Garg et al. \shortcite{garg2019jointly} leverage extracted discrete alignments in a multi-task framework to optimize towards translation and alignment objectives.

Inspired by the human learning process, curriculum learning~\cite{bengio2009curriculum} is proposed as a machine learning training strategy by feeding training instances to the model from easy to hard. Most of the works on curriculum learning focus on the determining the orders of data~\cite{lee2011learning,sachan2016easy}. Later, some works explore the curriculum learning strategies in task level. Previous work~\cite{sarafianos2017curriculum} in computer visual domain splits the tasks into groups according to the correlation and transfers the acquired knowledge from strongly correlated task to weakly correlated one.
%, while in our scenario, we find that the difficulties of tasks are related to $k$.

Guo et al. \shortcite{guo2019fine} propose a fine-tuning method to transfer a well-trained AT model to a NAT model by designing a curriculum in the shift process between two kinds of models, which is perhaps the most similar work to ours. However, the training strategy during their curriculum learning process is not a natural task, but just some hand-crafted training strategies, which could affect the final transfer accuracy and the total training time. In contrast, each intermediate task during our curriculum learning process is a standard translation task and is empirically verified to be helpful to the consequent tasks.

\section{Task-Level Curriculum Learning For NAT}
In this section, we introduce our proposed task-level curriculum learning for NAT (TCL-NAT) in detail. First, we propose a unified perspective to represent different tasks including AT, SAT and NAT with a parameter $k$. Second, we empirically demonstrate the task with smaller $k$ can help the task with bigger $k$. Third, we introduce the task-level curriculum learning mechanism based on the unified perspective. Finally, we describe the design of our model architecture for TCL-NAT.

\subsection{A Unified Perspective for AT/SAT/NAT}
\label{sec:perspective}
We propose a new perspective to view AT, SAT and NAT as to generate target tokens in an autoregressive manner during the whole sentence translation, but generate $k$ adjacent tokens in parallel at a time. Specifically, given a source and target sentence pair $(x, y) \in (\mathcal{X}, \mathcal{Y})$, we factorize the conditional probability $P(y|x)$ according to the chain rule:
\begin{equation}
\label{eq:cond_prob}
    P(y|x) = \prod_{t=0}^{\lfloor N/k \rfloor}  \prod_{j=1}^{k} P(y_{tk+j} | y_{< tk+1}, x; \theta), 
\end{equation}
where $N$ is the length of the target sequence, $k$ denotes the number of tokens that generated in parallel in a decoding step, $\lfloor \cdot \rfloor$ denotes the floor operation, $\theta$ denotes the parameters of the model. In the above equation, $y_{t}$ where $t<1$ or $t>N$ represents invalid tokens, which is introduced to make our formulation simple.

Under this perspective, we regard each $k$ as an individual task. As special cases, when $k=1$, the equation becomes:
\begin{equation}
    P(y|x) = \prod_{t=0}^{N}P(y_{t+1} | y_{< t+1}, x; \theta), 
\end{equation}
which is exactly the conditional probability for AT; when $k=N$, the equation becomes:
\begin{equation}
    P(y|x) = \prod_{j=1}^{N} P(y_{j}| x; \theta), 
\end{equation}
which is the conditional probability for NAT; when $1 < k < N$, the equation represents the conditional probability for SAT. 

\subsection{A Preliminary Study}
\label{sec:perspective}

Based on this perspective of $k$, we train multiple models\footnote{We use TCL-NAT model setting which is introduced in Section \ref{method_nat} for this preliminary study.} with task $k=1, 2, 4, 8, 16, N$ respectively and test them on the test set of IWSLT14 De-En dataset with different $k$. We have some analyses and observations:
\begin{itemize}
\item The task with smaller $k$ is easier to train and achieves higher accuracy, while that with larger $k$ is harder to train and achieves slightly worse accuracy~\cite{wang2018semi}, which forms a good curriculum way that shifts the task from easier to harder. The italic numbers in Table~\ref{tab_train_test_k_analysis} show that when training and testing the models with the same $k$, larger $k$ leads to worse accuracy. 
\item The model trained with task $k<N$ can bring advantages to the model training of task $k=N$, which can be supported in our experiments: we train multiple models with task $k=1, 2, 4, 8, 16$ respectively, and then test the translation accuracy of these models with task $k=N$ (NAT). We found most of the models (trained with $k=4,8,16$) can achieve reasonable accuracy on NAT, as shown in Table~\ref{tab_train_test_k_analysis}.
\item When testing the accuracy of task $k=N$, the model trained with task $k=k'<N$ brings more advantages than that trained with task $k<k'$. Similarly, when testing the accuracy of task $k=k'$, the model trained with task $k=k''<k'$ also provides a better initialization than that trained with task $k<k''$. Similar results can be got for smaller $k$ recursively. We can see from the bold numbers in Table~\ref{tab_train_test_k_analysis} that when testing the accuracy of task $k=N$, the model trained with $k=16$ achieves the best, when testing that of task $k=16$, the model trained with $k=8$ achieves the best and so on. 

\end{itemize}

\begin{table}[!h]
\centering
\small 
\begin{tabular}{p{0.9cm} p{0.6cm} p{0.6cm} p{0.6cm} p{0.6cm} p{0.6cm} p{0.7cm} }
\toprule
\diagbox[width=4em]{Test}{Train} & $k$=1 & $k$=2 & $k$=4 & $k$=8 & $k$=16 & $k$=N   \\ \midrule
    $k'$=N &0.28 & 1.35 & 6.39 & 19.38 & \textbf{23.78}  & \textit{24.89} \\
    $k'$=16 & 0.00 & 0.68 & 11.17 & \textbf{20.11} & \textit{24.97} & /      \\
    $k'$=8 & 0.17 & 1.54 & \textbf{12.87} & \textit{28.6} & / & /  \\
    $k'$=4 & 0.34 & \textbf{4.07} & \textit{31.27} & / & / & /   \\
    $k'$=2 & 0.86 & \textit{33.2} & / & / & /  & /  \\ 
    $k'$=1 & \textit{34.8} & / & / & / & /  & /             \\
\bottomrule
\end{tabular}
\caption{The BLEU scores on the test set of IWSLT14 De-En task. The model is trained with $k$ for 80k steps but test with another $k'$. The italic numbers show the accuracy of models that train and test with the same $k$. Row 1 shows that models trained with task $k=4,8,16$ can achieve reasonable accuracy on NAT. The bold numbers show that models trained with task $k=k''<k'$ can achieve better scores than that trained with task $k<k''$ when testing the accuracy of task $k=k'$.}
\label{tab_train_test_k_analysis}
\vspace{-0.2cm}
\end{table}

\subsection{Task-Level Curriculum Learning}
\label{sec:tcl}

Based on the unified perspective and the observations in the last subsections, we propose task-level curriculum learning for NAT (TCL-NAT). Our method consists of three phases: 
\begin{itemize}
    \item AT training. We first train a model with $k=1$ (AT) as the initial model. 
    \item SAT training. We then continue to train the model with increasing $k$ sequentially. For simplicity, we shift $k$ gradually and exponentially as $k=2, 4, 8, 16$ considering the length distribution of the dataset.
    \item NAT training. We continue to train the model with $k=N$ till convergence to obtain the final NAT model. 
\end{itemize}

Next, we elaborate the curriculum learning mechanism from two aspects: the pacing function for curriculum schedule and task window. Inspired by Guo et al. \shortcite{guo2019fine}, the pacing function is used to depict the curriculum scheduling strategy introduced in previous curriculum learning work, which controls the training step for each stage in SAT training phase. The task window represents the number of tasks trained at each stage, which is regarded as an extension to our method that only trains the model with one task at each stage.

\paragraph{Pacing Functions.} 
\label{para:pacing}
Specifically, at the $i$-th training step, we choose $k = f(i) \in \{2,4,8,16\}$, where $f(i)$ is the pacing function w.r.t the training step $i$. We define three different pacing functions: linear $f_{\textrm{linear}}(i)$, logarithmic $f_{\textrm{log}}(i)$ and exponential $f_{\textrm{exp}}(i)$ to make a smooth transformation from AT to NAT. These pacing functions divide the SAT training phase into several stages and there are clear differences among them: linear pacing function is the simplest pacing function where each stage is trained with same steps; logarithmic pacing function results in more training steps with tasks whose $k$ is larger and exponential pacing function shows more preference on tasks with smaller $k$. Since different pacing functions reflect different curriculum learning strategies, we conduct experiments to compare and analyze these proposed pacing functions. The detailed definitions and analysis of these pacing functions are shown in Section \ref{sec:analysis}.

\paragraph{Curriculum Learning with Task Window.} 
\label{para:cl_task_window}
If we just use the above mechanism for training, although the tasks between stages are similar and close, there still exists a gap between the tasks in neighboring stages. In order to reduce the gap between the stages, we try to extend the task-level curriculum learning with task window. Specifically, we define a task window $w$ to represent the number of tasks training at the same time at each stage. The situation we discussed before is corresponding to $w = 1$. When $w = 2$, we train the model with two tasks $k = 1,2$ at the same time in the first stage, then with two tasks $k = 2,4$ at the same time in the second stage, and so on until the last stage train with $k = 16,N$. When $w = 3$, our model is trained with three tasks $k = 1,2,4$ in the first stage, and so on. In this way, the training tasks in the two neighboring training stages have overlaps when $w>1$, which is smoother for task shift than $w=1$ intuitively.

\subsection{Model Structure for TCL-NAT}
\label{method_nat}

\begin{figure}[!ht]
	\centering
% 	\begin{subfigure}[h]{0.47\textwidth}
% 	\centering
% 	\includegraphics[width=\textwidth,trim={0cm 0.0cm 0cm 0cm}, clip=true]{fig/modela.pdf}
% 	\caption{TCL-NAT model  architecture.}
% 	\label{TCL-NAT model}
% 	\end{subfigure}
% 	\begin{subfigure}[h]{0.24\textwidth}
% 	\centering
% 	\includegraphics[width=\textwidth,trim={0cm 0.0cm 0cm 0cm}, clip=true]{fig/modelb.pdf}
% 	\caption{Encoder Block.}
% 	\label{encoder}
% 	\end{subfigure}
% 	\begin{subfigure}[h]{0.27\textwidth}
% 	\centering
% 	\includegraphics[width=\textwidth,trim={0cm 0.0cm 0cm 0cm}, clip=true]{fig/modelc.pdf}
% 	\caption{K-tokens Decoder Block.}
% 	\label{decoder_block}
% 	\end{subfigure}
% 	\begin{subfigure}[h]{0.2\textwidth}
% 	\centering
% 	\includegraphics[width=\textwidth,trim={0cm 0.0cm 0cm 0cm}, clip=true]{fig/modeld.pdf}
% 	\caption{K-tokens Mask.}
% 	\label{attn_mask}
% 	\end{subfigure}
	\includegraphics[width=0.48\textwidth,trim={0cm 0.0cm 0cm 0cm}, clip=true]{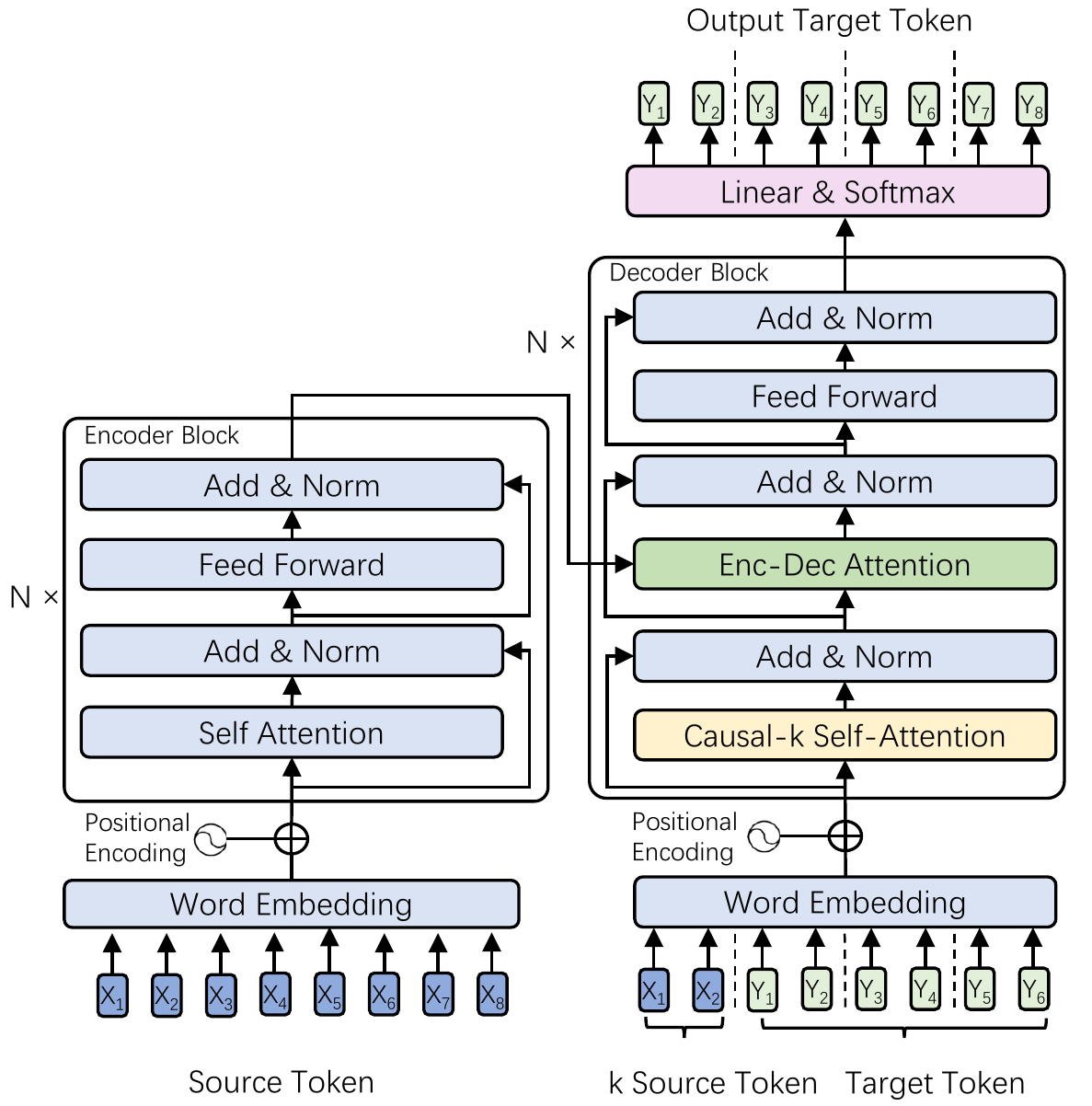}
	\caption{The overview of the model structure for TCL-NAT. This figure shows the case with $k=2$.}
	\label{model_fig}
	\vspace{-0.2cm}
\end{figure}

\begin{figure}[tb]
	\centering
	\includegraphics[width=0.45\textwidth,trim={0cm 0.0cm 0cm 0cm}, clip=true]{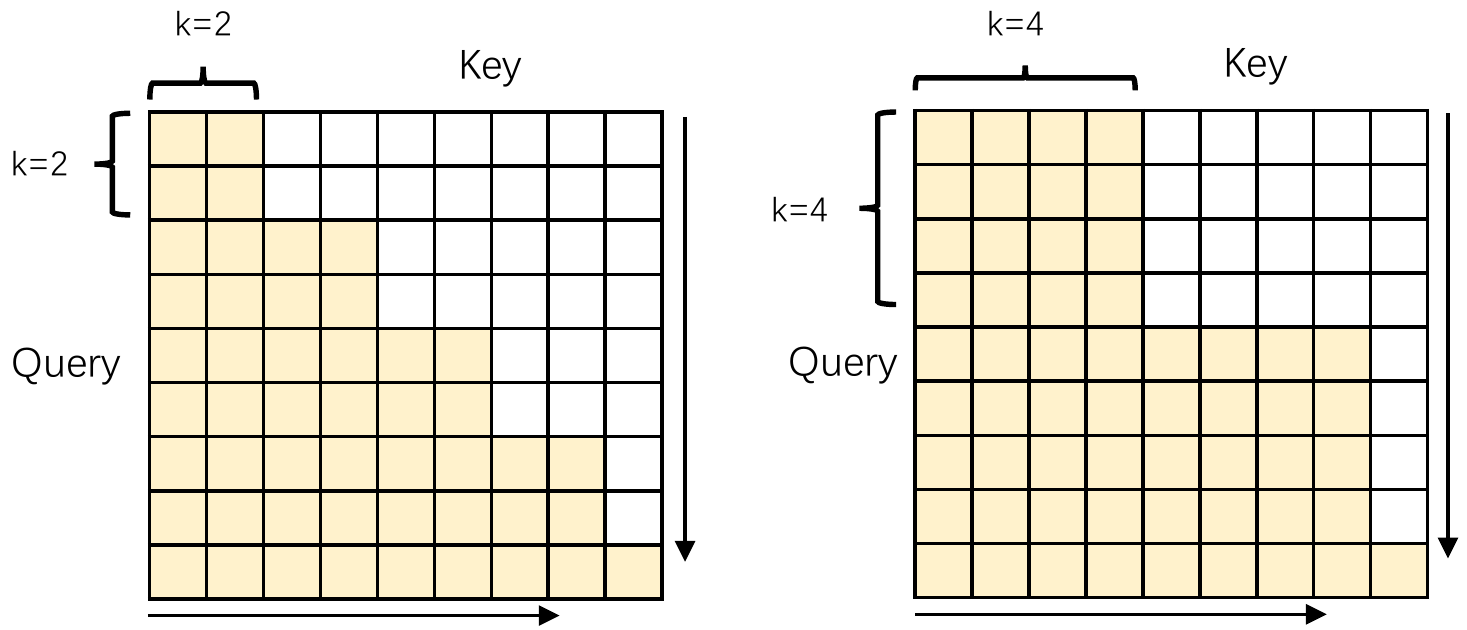}
	\caption{Causal-$k$ self-attention mask for an intuitive understanding. Yellow grids denote elements 1 and white grids denote elements 0 in the decoder self-attention mask. The left subfigure shows the case with $k=2$ and the right one shows the case with $k=4$.}
	\label{attn_mask}
	\vspace{-0.3cm}
\end{figure}

To support different $k$ in the same model, we leverage Transformer model with a causal-$k$ self-attention mechanism in the decoder~\cite{wang2018semi}. Note that although we only apply the task-level curriculum learning technique in Transformer-based model, it can also be easily applied to other non-autoregressive architectures such as CNN. The whole model architecture of TCL-NAT is shown in Figure \ref{model_fig}. The encoder of TCL-NAT is exactly the same as the basic structure of Transformer~\cite{DBLP:conf/nips/VaswaniSPUJGKP17}, which uses stacked self-attention and fully connected layers, shown in the left panel of the Figure \ref{model_fig}. For the decoder, we introduce a causal-$k$ self-attention mechanism which can generate $k$ successive tokens in parallel. As shown in the right panel of Figure \ref{model_fig}, our decoder is similar with the decoder in SAT \cite{wang2018semi}, except that we feed the model with the first $k$ source tokens $(x_1, ... x_k)$ rather than special tokens to predict $(y_1, ..., y_k)$ in parallel at the beginning of decoding, in order to keep consistent with NAT model when $k=N$. Then $(y_1, ..., y_k)$ are fed to the model to predict $(y_{k+1} ..., y_{2k})$ in parallel. As a result, the decoder input can be denoted as $ (x_1, ..., x_k, y_1, ... y_{N-k})$. We also adopt a causal-$k$ mask in the decoder self-attention following Wang et al. \shortcite{wang2018semi}, as shown in Figure \ref{attn_mask}. 

% Given target sentence length $N$ and $k$, the mask $Mask \in \mathbb{R}^{N \times N}$ can be defined as:
% \begin{equation}
%     Mask[i][j] = \left\{
%     \begin{array}{rcl}
%     1 &  {\text{if }  j < (\lfloor (i-1)/k \rfloor + 1) \times k }\\
%     0 &  {\text{other}}\\
%     \end{array} \right.,
% \end{equation}
% where $\lfloor \cdot \rfloor$ denotes the floor operation. In this way, model is allowed to access to $y_1 ... y_k$ rather than $y_1$ only when predicting $y_{k+1}$, 

Under this model structure, we can utilize the first $k$ source tokens $(x_1, ..., x_k)$ to predict sub-sentence $(y_1, ... y_k)$ in parallel, and then utilize them to predict sub-sentence $(y_{k+1}, ..., y_{2k})$ in parallel and so on. Then it comes the conditional probability in Equation \ref{eq:cond_prob}. As the $k$ increases, the model's dependency on target tokens decreases. Specially, when $k=1$ the decoder is an autoregressive decoder, and when $k$ is large enough, the decoder becomes a non-autoregressive decoder which generates all outputs simultaneously depending on the source tokens only. In the inference stage, we set $k$ to $N$ and make the decoder run in the NAT mode.

\section{Experiments and Results}
\subsection{Experiments Settings}
\paragraph{Datasets.}
We evaluate our method on three standard translation datasets: IWSLT14 German-to-English (De-En) dataset\footnote{https://wit3.fbk.eu/mt.php?release=2014-01}, IWSLT16 English-to-German (En-De) dataset\footnote{https://wit3.fbk.eu/mt.php?release=2016-01} and WMT14 English-to-German (En-De) dataset\footnote{https://www.statmt.org/wmt14/translation-task.html}. Following Li et al. \shortcite{li2019hint}, we reverse WMT14 English-to-German to get WMT14 German-to-English (De-En) dataset. In details, IWSLT14 dataset contains $153k/7k/7k$ parallel bilingual sentences for training/dev/test set respectively; IWSLT16 dataset contains $195k/1k/1k$ parallel bilingual sentences for training/dev/test set and WMT14 dataset contains 4.5M parallel sentence pairs for training sets, where newstest2014 and newstest2013 are used as test and validation set respectively, following previous works~\cite{gu2017non,guo2019fine}. We split each token into subwords using Byte-Pair Encoding (BPE)~\cite{sennrich2015neural} and set $10k$, $10k$ and $40k$ as the vocabulary size for IWSLT14, IWSLT16 and WMT14 respectively. The vocabulary is shared by source and target languages in those datasets.

\paragraph{Model Configuration.}
We adopt the basic NAT model configuration~\cite{gu2017non,guo2019fine} based on Transformer~\cite{DBLP:conf/nips/VaswaniSPUJGKP17}, which is composed by multi-head attention modules and feed forward networks. We follow Guo et al. \shortcite{guo2019fine} for configuration hyperparameters: For WMT14 datasets, we use the hyperparameters of a \texttt{base} Transformer ($d_{\textrm{model}}=d_{\textrm{hidden}}=512$, $n_{\textrm{layer}}=6$, $n_{\textrm{head}}=8$). For IWSLT14 and IWSLT16 datasets, we utilize a \texttt{small} Transformer ($d_{\textrm{model}}=d_{\textrm{hidden}}=256$, $n_{\textrm{layer}}=6$, $n_{\textrm{head}}=4$). 
% For  datasets, we use the settings ($d_{\textrm{model}}=287$, $d_{\textrm{hidden}}=507$, $n_{\textrm{layer}}=5$, $n_{\textrm{head}}=2$) following \cite{gu2017non}.

% Fairseq\footnote{https://github.com/pytorch/fairseq/blob/master/fairseq/models/\\transformer.py}~\cite{ott2019fairseq}

\paragraph{Training.}

Following previous works \cite{gu2017non,guo2019non,wang2019non}, we employ sequence-level knowledge distillation \cite{kim2016sequence} during training to reduce the difficulty of training and boost the accuracy through constructing a more deterministic and less noisy training set: first we train an AT teacher model which has the same architecture as the TCL-NAT model, and then we use the translation results of each source sentence generated by the teacher model as the new ground truth to construct a new training set for further training. We design three different pacing functions mentioned in Section \ref{para:cl_task_window} and detail their definitions in Table \ref{tab_cl_functions}. We set task window $w$ to 2 by default, which is determined by the model performance on the validation sets. We train all models using Adam following the optimizer settings and learning rate schedule in Transformer \cite{DBLP:conf/nips/VaswaniSPUJGKP17}. We run the training procedure on 8 NVIDIA Tesla P100 GPUs for WMT and 2 NVIDIA 2080Ti GPUs for IWSLT datasets respectively. The training steps of each phase are listed in Table \ref{tab:steps}. We implement our model on Tensor2Tensor \cite{vaswani2018tensor2tensor}.

\newcommand{\tabincell}[2]{\begin{tabular}{@{}#1@{}}#2\end{tabular}}
\begin{table}[!h]
\centering
\begin{tabular}{l | c c c c}
\toprule
\textbf{Phases} & \tabincell{c}{IWSLT14\\De-En} & \tabincell{c}{IWSLT16\\En-De} & \tabincell{c}{WMT14\\De-En} &  \tabincell{c}{WMT14\\En-De}    \\ 
\midrule
AT           & 0.08M & 0.08M  & 0.16M  & 0.16M  \\
SAT       & 0.24M & 0.24M  & 0.32M  & 0.32M  \\
NAT           & 0.32M & 0.32M  & 0.64M  & 0.64M  \\
\bottomrule
\end{tabular}
\caption{The training steps of TCL-NAT for different datasets for each phase.}
\label{tab:steps}
\vspace{-0.3cm}
\end{table}

\begin{table}[!h]
\centering
\begin{tabular}{l l}
\toprule
Name & Description \\\midrule

Linear  & 

{
% \begin{equation*}
 $f_{\textrm{linear}}(i) = \min(2^{\lfloor 4i/S_{\textrm{SAT}}  \rfloor  + 1}, 16)$
% \end{equation*}
}

\\
\midrule
Logarithmic &
{
% \begin{equation*}
 $f_{\textrm{log}}(i) = \min(2^{\lfloor log_{1.5}(4i/S_{\textrm{SAT}}+1) \rfloor +1 } , 16)$
% \end{equation*}
}

\\\midrule

Exponential & 

{
% \begin{equation*}
 $f_{\textrm{exp}}(i) = \min(2^{\lfloor 1.5^{4i/S_{\textrm{SAT}}}  \rfloor}, 16)$
% \end{equation*}
}

\\
\bottomrule
\end{tabular}
\caption{The proposed different curriculum pacing functions and their definitions. $S_{\textrm{SAT}}$ denotes the total steps in SAT training phase. We choose constants empirically to meet the actual training situation.}
\label{tab_cl_functions}
\vspace{-0.3cm}
\end{table}

\paragraph{Inference and Evaluation.}
For inference, we adopt the common method of noisy parallel decoding (NPD) \cite{gu2017non}, which generates a number of decoding candidates in parallel and selects the best translation by AT teacher model re-scoring. In our work, we generate multiple translation candidates by predicting different target lengths $N \in [M - B, M + B]$ ($M$ is the length of the source sentence), which results in $2B+1$ candidates.  We test with $B = 0$ and $B = 4$ (denoted as NPD 9) to keep consistent with our baselines \cite{wang2019non,guo2019non,guo2019fine}.
We evaluate the translation quality by tokenized case sensitive BLEU~\cite{DBLP:conf/acl/PapineniRWZ02} with multi-bleu.pl\footnote{https://github.com/moses-smt/mosesdecoder/blob \\ /master/scripts/generic/multi-bleu.perl}. Inference and evaluation are run on $1$ Nvidia P100 GPU for WMT14 En-De datasets in order to keep in line with previous works for testing latency.
% In our experiments, we set $\Delta T$ to $0, 0, -1, 1$ for IWSLT14 De-En, IWSLT16 En-De, WMT14 De-En and WMT14 En-De respectively according to the average length distribution in the training set.

\begin{table*}[!ht]
    \centering
    \begin{tabular}{l | c c c c | c c}
    \toprule
    \textbf{Models} & \tabincell{c}{IWSLT14\\De-En} & \tabincell{c}{IWSLT16\\En-De} & \tabincell{c}{WMT14\\De-En} &  \tabincell{c}{WMT14\\En-De} & Latency & Speedup          \\ 
    \midrule
    \multicolumn{1}{l}{\textit{Autoregressive Models (AT Teachers)}}         \\
    \midrule
    Transformer~\cite{DBLP:conf/nips/VaswaniSPUJGKP17}       & 33.90 & 30.32 & 31.38 & 27.30  & 607 ms & 1.00 $\times$    \\
    \midrule
    \multicolumn{1}{l}{\textit{Non-Autoregressive Models}}         \\
    \midrule
    NAT-FT \cite{gu2017non}   & / & 26.52 & 21.47 & 17.69  & 39 ms  & 15.6 $\times$     \\
    NAT-FT (NPD 10)   & /  & 27.44 & 22.41  & 18.66      & 79 ms  & 7.68 $\times$     \\
    NAT-IR \cite{lee2018deterministic}   & /  & 27.11 & 25.48  & 21.61   & 404 ms  & 1.50 $\times$     \\
    ENAT \cite{guo2019non}   & 25.09  & / & 23.23  & 20.65 & 24 ms  & 25.3 $\times$     \\
    ENAT (NPD 9)   & 28.60  & / & 26.67  & 24.28 & 49 ms  & 12.4 $\times$     \\
    NAT-Reg \cite{wang2019non} & 23.89  & 23.14 & 24.77  & 20.65 & 22 ms  & 27.6 $\times$     \\
    NAT-Reg (NPD 9) & 28.04  & 27.02 & 28.90  & 24.61 & 40 ms  & 15.1 $\times$     \\
    FlowSeq-base \cite{ma2019flowseq} & 27.55 & / & 26.16 & 21.45 & / & 5.94 $\times$ \\
    FCL-NAT \cite{guo2019fine}  & 26.62 & / & 25.32 & 21.70 & 21 ms  & 28.9 $\times$     \\
    FCL-NAT (NPD 9)  & 29.91 & / & 29.50 & \textbf{25.75} & 38 ms  & 16.0 $\times$     \\
    \midrule
    TCL-NAT      & 28.16 & 26.01 & 25.62   & 21.94    & 22 ms  & 27.6 $\times$              \\
    TCL-NAT (NPD 9)     & \textbf{31.79} & \textbf{29.30}  & \textbf{29.60}   & 25.37  & 38 ms  & 16.0 $\times$ \\

\bottomrule
\end{tabular}
\caption{The BLEU scores of our proposed TCL-NAT and the baseline methods on the IWSLT14 De-En, IWSLT16 En-De, WMT14 De-En and WMT14 En-De tasks. NPD 9 indicates results of noisy parallel decoding with 9 candidates, i.e., B = 4, otherwise B = 0. }
\label{tab:main_res}
\vspace{-0.2cm}
\end{table*}

\subsection{Results}

We compare TCL-NAT with non-autoregressive baselines including NAT-FT~\cite{gu2017non}, NAT-IR~\cite{lee2018deterministic}, ENAT~\cite{guo2019non}, NAT-Reg~\cite{wang2019non}, FlowSeq~\cite{ma2019flowseq} and FCL-NAT~\cite{guo2019fine}. For NAT-IR, we report their best results when refinement steps is 10. For ENAT, NAT-Reg and FCL-NAT, we report their best results with B = 0 and B = 4 correspondingly. For FlowSeq, we report their results without NPD. It is worth noting that we mainly compare our method with existing methods that have similar speed-up, so Mask-Predict~\cite{ghazvininejad2019mask}, LevT~\cite{gu2019levenshtein} and FlowSeq-large are not included into discussion.

We list the main results of our work in Table \ref{tab:main_res}. We can see that TCL-NAT achieves significant improvements over all NAT baselines on different datasets. Specifically, we outperform ENAT and NAT-Reg with a notable margin. In addition, compared with NAT-Reg, we do not introduce any auxiliary loss functions in training stage and compared with ENAT, we just copy the source sentence as the decoder input, which does not add the extra workload in inference stage. Compared with FlowSeq, our method (without NPD) achieves better scores on most datasets with a much larger speedup. We also outperform FCL-NAT on most datasets with a less training steps. As for the inference efficiency, we achieve a 16.0 times speedup (NPD 9), which is comparable with state of the art methods (FCL-NAT and ENAT).

\subsection{Analyses}
\label{sec:analysis}

\paragraph{Comparison with Direct Transfer.}
We take Direct Transfer (DT) as another baseline, where we omit the SAT stage in Section \ref{sec:tcl}, and train the model in a non-autoregressive manner for the same steps as TCL-NAT to ensure a fair comparison. We test DT model on the test set of IWSLT14 De-En task and obtain the BLEU score of 27.00, while our method achieves 28.16 BLEU score. We can see that compared with DT, TCL-NAT gains a large improvement on translation accuracy, demonstrating the importance of the progressive transfer between two tasks with curriculum learning.

\paragraph{Analysis on Pacing Functions.}
We compare the accuracy of models trained with different pacing functions. We evaluate TCL-NAT with different pacing functions shown in Table \ref{tab_cl_functions}. From the table, we can see that the model trained with exponential function slightly outperforms those with other functions and the logarithmic function performs the worst. As we mentioned in Section \ref{para:pacing}, exponential function shows more preference on easier stages while logarithmic function focuses more on harder stages, and therefore we can conclude that showing more preference on easier tasks is beneficial to the NAT model training, and thus obtain a better score.

\begin{table}[!ht]
\centering
\begin{tabular}{l | c c c}
\toprule
\textbf{Pacing Functions} & Linear & Logarithmic & Exponential \\ 
\midrule
TCL-NAT      & 27.89  & 27.76  & \textbf{27.96}  \\
TCL-NAT (NPD 9)  & 31.51  & 31.45   & \textbf{31.71}  \\
\bottomrule
\end{tabular}
\caption{The comparison of BLEU scores on the test set of IWSLT14 De-En task among different pacing functions.}
\label{tab:pacing_funcs}
\vspace{-0.2cm}
\end{table}

\paragraph{Analysis on Task Window.}
We compare the accuracy of models trained with different task windows, as mentioned in Section \ref{para:cl_task_window}. The results are listed in Table \ref{tab:window_size}. From the table, we can see that the model trained with $w=2$ achieves the best score in IWSLT14 De-En task, which proves that an appropriate task window $w$ can help reduce the gap between neighboring stages, and thus help model training.

\begin{table}[!ht]
\centering
\begin{tabular}{l | c c c c }
\toprule
\textbf{Task Window} & $w=1$ & $w=2$ & $w=3$ & $w=4$  \\ 
\midrule
TCL-NAT      & 27.89  & \textbf{28.16}   & 28.00 & 27.96 \\
TCL-NAT (NPD 9)  & 31.51  & \textbf{31.79}   & 31.44 & 31.40  \\
\bottomrule
\end{tabular}
\caption{The comparison of BLEU scores on the test set of IWSLT14 De-En task among different task windows.}
\label{tab:window_size}
\vspace{-0.2cm}
\end{table}

% \begin{table}[!ht]
% \centering
% \begin{tabular}{l | c c}
% \toprule
% \textbf{Models} & BLEU  \\ 
% \midrule
% DT      & 27.00   \\
% TCL-NAT      & 28.16  \\  
% \bottomrule
% \end{tabular}
% \caption{The comparison of BLEU scores on the test set of IWSLT14 De-En task between DT (Direct Transfer) and TCL-NAT. }
% \label{tab:dt}
% \end{table}

\section{Conclusion}
In this work, we proposed a novel task-level curriculum learning method to improve the accuracy of non-autoregressive neural machine translation. We first view autoregressive, semi-autoregressive and non-autoregressive translation as individual tasks with different $k$, and propose a task-level curriculum mechanism to shift the training process from $k=1$ to $N$, where $N$ is the length of the target sentence. Experiments on several benchmark translation datasets demonstrate the effectiveness of our method for NAT.

In the future, we will extend the task-level curriculum learning method to other sequence generation tasks such as non-autoregressive speech synthesis, automatic speech recognition and image captioning, where there exists smooth transformation between autoregressive  and non-autoregressive generation using semi-autoregressive generation as bridges. We expect task-level curriculum learning could become a general training paradigm for a broader range of tasks.

\section*{Acknowledgments}
This work was supported in part by the National Key R\&D Program of China (Grant No.2018AAA0100603), Zhejiang Natural Science Foundation (LR19F020006), National Natural Science Foundation of China (Grant No.61836002), National Natural Science Foundation of China (Grant No.U1611461), National Natural Science Foundation of China (Grant No.61751209), and Microsoft Research Asia.

\newpage

\bibliographystyle{named}
\bibliography{ijcai20}

\begin{thebibliography}{}

\bibitem[\protect\citeauthoryear{Anastasopoulos and
  Chiang}{2018}]{anastasopoulos2018tied}
Antonios Anastasopoulos and David Chiang.
\newblock Tied multitask learning for neural speech translation.
\newblock In {\em NAACL}, pages 82--91, June 2018.

\bibitem[\protect\citeauthoryear{Bahdanau \bgroup \em et al.\egroup
  }{2015}]{DBLP:journals/corr/BahdanauCB14}
Dzmitry Bahdanau, Kyunghyun Cho, and Yoshua Bengio.
\newblock Neural machine translation by jointly learning to align and
  translate.
\newblock {\em ICLR}, 2015.

\bibitem[\protect\citeauthoryear{Bengio \bgroup \em et al.\egroup
  }{2009}]{bengio2009curriculum}
Yoshua Bengio, J{\'e}r{\^o}me Louradour, Ronan Collobert, and Jason Weston.
\newblock Curriculum learning.
\newblock In {\em ICML}, pages 41--48. ACM, 2009.

\bibitem[\protect\citeauthoryear{Caruana}{1997}]{caruana1997multitask}
Rich Caruana.
\newblock Multitask learning.
\newblock {\em Machine learning}, 28(1):41--75, 1997.

\bibitem[\protect\citeauthoryear{Devlin \bgroup \em et al.\egroup
  }{2019}]{devlin2018bert}
Jacob Devlin, Ming-Wei Chang, Kenton Lee, and Kristina Toutanova.
\newblock Bert: Pre-training of deep bidirectional transformers for language
  understanding.
\newblock In {\em NAACL-HLT}, 2019.

\bibitem[\protect\citeauthoryear{Dong \bgroup \em et al.\egroup
  }{2015}]{dong2015multi}
Daxiang Dong, Hua Wu, Wei He, Dianhai Yu, and Haifeng Wang.
\newblock Multi-task learning for multiple language translation.
\newblock In {\em ACL-IJCNLP}, pages 1723--1732, 2015.

\bibitem[\protect\citeauthoryear{Garg \bgroup \em et al.\egroup
  }{2019}]{garg2019jointly}
Sarthak Garg, Stephan Peitz, Udhyakumar Nallasamy, and Matthias Paulik.
\newblock Jointly learning to align and translate with transformer models.
\newblock In {\em EMNLP-IJCNLP}, pages 4452--4461, November 2019.

\bibitem[\protect\citeauthoryear{Gehring \bgroup \em et al.\egroup
  }{2017}]{DBLP:conf/icml/GehringAGYD17}
Jonas Gehring, Michael Auli, David Grangier, Denis Yarats, and Yann~N. Dauphin.
\newblock Convolutional sequence to sequence learning.
\newblock In {\em ICML}, pages 1243--1252, 2017.

\bibitem[\protect\citeauthoryear{Ghazvininejad \bgroup \em et al.\egroup
  }{2019}]{ghazvininejad2019mask}
Marjan Ghazvininejad, Omer Levy, Yinhan Liu, and Luke Zettlemoyer.
\newblock Mask-predict: Parallel decoding of conditional masked language
  models.
\newblock In {\em EMNLP-IJCNLP}, pages 6114--6123, 2019.

\bibitem[\protect\citeauthoryear{Gu \bgroup \em et al.\egroup
  }{2018}]{gu2017non}
Jiatao Gu, James Bradbury, Caiming Xiong, Victor~O.K. Li, and Richard Socher.
\newblock Non-autoregressive neural machine translation.
\newblock In {\em ICLR}, 2018.

\bibitem[\protect\citeauthoryear{Gu \bgroup \em et al.\egroup
  }{2019}]{gu2019levenshtein}
Jiatao Gu, Changhan Wang, and Junbo Zhao.
\newblock Levenshtein transformer.
\newblock In {\em Advances in Neural Information Processing Systems}, pages
  11179--11189, 2019.

\bibitem[\protect\citeauthoryear{Guo \bgroup \em et al.\egroup
  }{2019a}]{guo2019non}
Junliang Guo, Xu~Tan, Di~He, Tao Qin, Linli Xu, and Tie-Yan Liu.
\newblock Non-autoregressive neural machine translation with enhanced decoder
  input.
\newblock In {\em AAAI}, volume~33, pages 3723--3730, 2019.

\bibitem[\protect\citeauthoryear{Guo \bgroup \em et al.\egroup
  }{2019b}]{guo2019fine}
Junliang Guo, Xu~Tan, Linli Xu, Tao Qin, Enhong Chen, and Tie-Yan Liu.
\newblock Fine-tuning by curriculum learning for non-autoregressive neural
  machine translation.
\newblock {\em arXiv preprint arXiv:1911.08717}, 2019.

\bibitem[\protect\citeauthoryear{Kim and Rush}{2016}]{kim2016sequence}
Yoon Kim and Alexander~M. Rush.
\newblock Sequence-level knowledge distillation.
\newblock In {\em EMNLP}, pages 1317--1327, 2016.

\bibitem[\protect\citeauthoryear{Lee and Grauman}{2011}]{lee2011learning}
Yong~Jae Lee and Kristen Grauman.
\newblock Learning the easy things first: Self-paced visual category discovery.
\newblock In {\em CVPR 2011}, pages 1721--1728. IEEE, 2011.

\bibitem[\protect\citeauthoryear{Lee \bgroup \em et al.\egroup
  }{2018}]{lee2018deterministic}
Jason Lee, Elman Mansimov, and Kyunghyun Cho.
\newblock Deterministic non-autoregressive neural sequence modeling by
  iterative refinement.
\newblock In {\em EMNLP}, pages 1173--1182, 2018.

\bibitem[\protect\citeauthoryear{Li \bgroup \em et al.\egroup
  }{2019}]{li2019hint}
Zhuohan Li, Zi~Lin, Di~He, Fei Tian, QIN Tao, WANG Liwei, and Tie-Yan Liu.
\newblock Hint-based training for non-autoregressive machine translation.
\newblock In {\em EMNLP-IJCNLP}, pages 5712--5717, 2019.

\bibitem[\protect\citeauthoryear{Ma \bgroup \em et al.\egroup
  }{2019}]{ma2019flowseq}
Xuezhe Ma, Chunting Zhou, Xian Li, Graham Neubig, and Eduard Hovy.
\newblock Flowseq: Non-autoregressive conditional sequence generation with
  generative flow.
\newblock In {\em EMNLP-IJCNLP}, pages 4273--4283, 2019.

\bibitem[\protect\citeauthoryear{Papineni \bgroup \em et al.\egroup
  }{2002}]{DBLP:conf/acl/PapineniRWZ02}
Kishore Papineni, Salim Roukos, Todd Ward, and Wei{-}Jing Zhu.
\newblock Bleu: a method for automatic evaluation of machine translation.
\newblock In {\em ACL}, pages 311--318, 2002.

\bibitem[\protect\citeauthoryear{Ren \bgroup \em et al.\egroup
  }{2019}]{ren2019fastspeech}
Yi~Ren, Yangjun Ruan, Xu~Tan, Tao Qin, Sheng Zhao, Zhou Zhao, and Tie-Yan Liu.
\newblock Fastspeech: Fast, robust and controllable text to speech.
\newblock {\em arXiv preprint arXiv:1905.09263}, 2019.

\bibitem[\protect\citeauthoryear{Sachan and Xing}{2016}]{sachan2016easy}
Mrinmaya Sachan and Eric Xing.
\newblock Easy questions first? a case study on curriculum learning for
  question answering.
\newblock In {\em ACL}, volume~1, pages 453--463, 2016.

\bibitem[\protect\citeauthoryear{Sarafianos \bgroup \em et al.\egroup
  }{2017}]{sarafianos2017curriculum}
Nikolaos Sarafianos, Theodore Giannakopoulos, Christophoros Nikou, and
  Ioannis~A Kakadiaris.
\newblock Curriculum learning for multi-task classification of visual
  attributes.
\newblock In {\em ICCV}, pages 2608--2615, 2017.

\bibitem[\protect\citeauthoryear{Sennrich \bgroup \em et al.\egroup
  }{2016}]{sennrich2015neural}
Rico Sennrich, Barry Haddow, and Alexandra Birch.
\newblock Neural machine translation of rare words with subword units.
\newblock In {\em ACL}, pages 1715--1725, 2016.

\bibitem[\protect\citeauthoryear{Song \bgroup \em et al.\egroup
  }{2019}]{song2019mass}
Kaitao Song, Xu~Tan, Tao Qin, Jianfeng Lu, and Tie-Yan Liu.
\newblock Mass: Masked sequence to sequence pre-training for language
  generation.
\newblock In {\em ICML}, pages 5926--5936, 2019.

\bibitem[\protect\citeauthoryear{Sutskever \bgroup \em et al.\egroup
  }{2014}]{sutskever2014sequence}
I~Sutskever, O~Vinyals, and QV~Le.
\newblock Sequence to sequence learning with neural networks.
\newblock {\em Advances in NIPS}, 2014.

\bibitem[\protect\citeauthoryear{Vaswani \bgroup \em et al.\egroup
  }{2017}]{DBLP:conf/nips/VaswaniSPUJGKP17}
Ashish Vaswani, Noam Shazeer, Niki Parmar, Jakob Uszkoreit, Llion Jones,
  Aidan~N. Gomez, Lukasz Kaiser, and Illia Polosukhin.
\newblock Attention is all you need.
\newblock In {\em NIPS}, pages 6000--6010, 2017.

\bibitem[\protect\citeauthoryear{Vaswani \bgroup \em et al.\egroup
  }{2018}]{vaswani2018tensor2tensor}
Ashish Vaswani, Samy Bengio, Eugene Brevdo, Francois Chollet, Aidan Gomez,
  Stephan Gouws, Llion Jones, {\L}ukasz Kaiser, Nal Kalchbrenner, Niki Parmar,
  et~al.
\newblock Tensor2tensor for neural machine translation.
\newblock In {\em AMTA}, pages 193--199, 2018.

\bibitem[\protect\citeauthoryear{Wang \bgroup \em et al.\egroup
  }{2018}]{wang2018semi}
Chunqi Wang, Ji~Zhang, and Haiqing Chen.
\newblock Semi-autoregressive neural machine translation.
\newblock In {\em EMNLP}, pages 479--488, 2018.

\bibitem[\protect\citeauthoryear{Wang \bgroup \em et al.\egroup
  }{2019}]{wang2019non}
Yiren Wang, Fei Tian, Di~He, Tao Qin, ChengXiang Zhai, and Tie-Yan Liu.
\newblock Non-autoregressive machine translation with auxiliary regularization.
\newblock In {\em AAAI}, 2019.

\end{thebibliography}

\end{document}